\documentclass[conference]{IEEEtran}
\IEEEoverridecommandlockouts
\usepackage{cite}
\usepackage{amsmath,amssymb,amsfonts}
\usepackage{xspace}
\usepackage{algorithmic}
\usepackage{graphicx}
\usepackage{textcomp}
\usepackage{lmodern}
\usepackage[T1]{fontenc}
\usepackage{xcolor}
\usepackage{tabularx}
\usepackage{caption}
\usepackage{subcaption}
\usepackage{comment}
\usepackage[bookmarks=false]{hyperref}
\usepackage{float}
\usepackage{amssymb}
\def\BibTeX{{\rm B\kern-.05em{\sc i\kern-.025em b}\kern-.08em
    T\kern-.1667em\lower.7ex\hbox{E}\kern-.125emX}}
\begin{document}

\title{Question Answering over Knowledge Base using Language Model Embeddings}

\author{\IEEEauthorblockN{ Sai Sharath, Japa}
\IEEEauthorblockA{\textit{School of Computing} \\
\textit{Southern Illinois University}\\
Carbondale, IL \\
sharath.japa@siu.edu}
\and
\IEEEauthorblockN{ Banafsheh, Rekabdar}
\IEEEauthorblockA{\textit{School of Computing} \\
\textit{Southern Illinois University}\\
Carbondale, IL \\
brekabdar@cs.siu.edu}}

\maketitle

\begin{abstract}
Knowledge Base, represents facts about the world, often in some form of subsumption ontology, rather than implicitly, embedded in procedural code, the way a conventional computer program does. While there is a rapid growth in knowledge bases, it poses a challenge of retrieving information from them. Knowledge Base Question Answering is one of the promising approaches for extracting substantial knowledge from Knowledge Bases. Unlike web search, Question Answering over a knowledge base gives accurate and concise results, provided that natural language questions can be understood and mapped precisely to an answer in the knowledge base. However, some of the existing embedding-based methods for knowledge base question answering systems ignore the subtle correlation between the question and the Knowledge Base (e.g., entity types, relation paths, and context) and suffer from the Out Of Vocabulary problem. In this paper, we focused on using a pre-trained language model for the Knowledge Base Question Answering task. Firstly, we used Bert base uncased for the initial experiments. We further fine-tuned these embeddings with a two-way attention mechanism from the knowledge base to the asked question and from the asked question to the knowledge base answer aspects. Our method is based on a simple Convolutional Neural Network architecture with a Multi-Head Attention mechanism to represent the asked question dynamically in multiple aspects. Our experimental results show the effectiveness and the superiority of the Bert pre-trained language model embeddings for question answering systems on knowledge bases over other well-known embedding methods. 



\end{abstract}


\section{Introduction}
Question Answering (QA) systems enable natural language platforms to interact with the Knowledge Bases. These QA systems return direct and more specific answers to the asked questions. In recent years, with the increase in the popularity and the use of knowledge bases (KB) such as Google's Knowledge Graph \cite{singhal2012introducing}, YAGO \cite{10.1145/1242572.1242667}, DBPedia \cite{lehmann2015dbpedia}, and Freebase\cite{bollacker2008freebase} people are more interested in seeking effective methods to access these knowledge bases. Most of such knowledge bases adopt Resource Description Framework (RDF) as their data format and they also contain billions of SPO (subject, predicate, object) triples\cite{klyne2009resource}. 

There are several languages designed for querying such large KBs, including SPARQL\cite{seaborne2008sparql}, Xcerpt\cite{10.1007/978-3-540-27775-0_33}, RQL\cite{inproceedings}. However, learning these languages adds a limitation, as one needs to be familiarized with the query language, it's syntax, it's semantics, and the ontology of the knowledge base. 

By contrast, KB-QA which takes natural language as it's query is a more user-friendly solution and became a research focus in recent years \cite{unger2014introduction}. There are two main research streams for the task of QA: Semantic Parsing-based systems (SP-based) \cite{zettlemoyer2009learning}, \cite{zettlemoyer2012learning,cui2017kbqa}, and Information Retrieval-based systems (IR-based) \cite{yao2014information, hao2017end},\cite{ bordes2014question, bordes2015large}. SP-based methods address the QA problem by constructing a semantic parser that converts the natural language question into a conditionally structured expressions like the logical forms\cite{alshawi1989logical} and then run the query on the Knowledge base to obtain the answer. SP- based methods consists of three modules: (1) Entity linking, recognizes all entity mentions in a question  and links each mention to an entity in KB; (2)Predicate mapping, finds candidate predicates in KB for the question;(3) Answer selection, converts the candidate entity-predicate pairs into a query statement and queries the knowledge base to obtain the answer. IR-based methods\cite{yao2014information, hao2017end},\cite{ bordes2014question, bordes2015large} focus on mapping answers and questions into the same embedding space, where one could query any KB, independent of its schema without requiring any grammar or lexicon. IR-based methods are more flexible and require less supervision compared with the SP-based approaches \cite{zettlemoyer2009learning},\cite{zettlemoyer2012learning,cui2017kbqa}. 

With the advancement in the embedding techniques which can capture the semantics, deep neural networks are applied in several areas of Natural Language Processing (NLP) \cite{10.1007/978-3-642-77189-7_5} and Natural Language Understanding (NLU) \cite{winograd1983language}. In the field of KB-QA, under IR-based umbrella, embedding-based approaches \cite{bordes2014open} \cite{hao-etal-2017-end}have been proposed. However, these approaches face two limitations. First, these models encode different components separately without learning the representation of the whole KB. Hence, are not able to capture the compositional semantics in a global perspective. Second, the performance of Long Short Term Memory\cite{sak2014long}(LSTM)s, Bi-directional LSTM\cite{10.5555/1986079.1986220}(BiLSTM)s, and Convolutional Neural Network\cite{fukushima1988neocognitron}(CNN)s are heavily dependent on large training data sets which is often not available in practice. In recent years, pre-trained Language Models\cite{devlin2018bert}\cite{radford2019language}\cite{radford2018improving}\cite{peters2018deep} on large-scale unsupervised corpus has shown its advantages on mining prior linguistic knowledge automatically, it indicates a possible way to deal with above problems. 

In comparison to previous embedding-based approaches, we focus on exploiting pre-trained language models for IR based KB-QA task. We use BERT \cite{devlin2018bert} pre-trained language model embeddings to encode our question and candidate answer contexts. 
We exploit a multi head attention encoder based on Convolution Neural Network (CNN) encoder \cite{DBLP:journals/corr/VaswaniSPUJGKP17} \cite{kim2014convolutional}  to fine tune Bert pre-trained embeddings in particular for the KB-QA task. Since we use a Language Model trained on 
the English Wikipedia and Brown Corpus \cite{devlin2018bert} for our question and KB representations, Out Of Vocabulary (OOV) problem can be negated. The interrelationships between the questions and the underlying KB are stored as 
the context for our model. This context is used to filter out the candidate answers by 
implementing cross attention between the question being asked and the KB.

The contribution of this paper is summarized as follows:
(1) 
We propose a method called \textit {Language Model based Knowledge Base Question Answering} (LM-KBQA). It exploits the BERT pre-trained language model embeddings \cite{devlin2018bert}
( which eliminates the need for using Recurrent Neural Network(RNN) architectures (LSTM, Bi-LSTM) to capture the contextual representations); (2) it is based on CNN encoder with self multi-head attention mechanism to fine tune the Bert embedding for the KB-QA task; (3) 
our results demonstrate the effectiveness of the proposed approach on the open Web-Questions data-set \cite{berant-etal-2013-semantic}.

\section{Related Work}

Over the past few years we have seen a growing research on KB-QA, shaping interactive paradigm. This enables user to take advantage of the communicative power of linguistic web knowledge while  hiding their complexity behind an easy-to-use interface. At the same time the abundance of information has led to a heterogeneous data landscape where QA systems struggle to stay up with the quantity, variety, and truthfulness of the underlying knowledge. In general, the most popular methods for KB-QA can be mainly divided into two classes: SP-based and IR-based.

Semantic parsing(SP) based approaches focuses on constructing a semantic parsing tree or equivalent query structure that represents the semantic meaning of the question. In terms of logical representation of natural language questions, many methods have been implemented, such as query graph \cite{yih-etal-2014-semantic,yih-etal-2015-semantic} or RDF query language \cite{cui2017kbqa}\cite{hu2017answering}.

Information retrieval(IR) based system try to obtain target answer directly from question information and KB knowledge without explicit considering interior query structure. There are various methods \cite{yao2014information, bordes2015large, dong-etal-2015-question,xu-etal-2016-question} to select candidate answers and rank results.

\cite{bordes2014open} Were the pioneers to use embedding-based models to resolve the KB-QA problem. The queries and KB triples were represented as vectors in an exceedingly low dimensional vector space. Subsequently the cosine similarity is applied to find the most accurate answer. Also the Bag Of Words (BOW) methodology \cite{harris1954distributional} is used to generate one vector for all the query answers. The Pairwise training methodology is utilized and the negative samples are randomly extracted from the KB. Authors in \cite{bordes2014question} improved their work based on sub-graph embeddings which is capable of answering complex questions. Their main motive was to include the maximum amount of information within the answer context which encodes the surrounding sub-graph of the KB(e.g., answer path and context). The proposed sub-graph embeddings accommodate all the entities and relations that are connected to the answer entity. The resulting vector was obtained using BOW\cite{harris1954distributional} representations. In follow-up work \cite{bordes2015large}\cite{jain2016question}, memory networks \cite{weston2014memory} were used to store candidate set, and could be iteratively accessed to mimic multi-hop reasoning. Unlike the above methods that mainly use a bag-of-words (BOW) approach to encode questions and KB contexts, \cite{dong-etal-2015-question,hao-etal-2017-end} apply more advanced network modules (e.g., CNNs and LSTMs). The approach proposed in \cite{yih-etal-2014-semantic} is fixated on single-relation queries. The KB-QA task was sub-divided into 2 stages. First and foremost, the topic entity of the question was found. Then, the later question was considered by CNNs and utilize to match relations. Authors in \cite{dong-etal-2015-question}, thought of the diverse features of answers to represent questions respectively using 3 columns of CNNs. Authors in \cite{xu-etal-2016-hybrid,xu-etal-2016-question} proposed to use multi-channel CNNs to extract the relations of KB triples while exploiting the free text of the Wikipedia.

Most embedding-based approaches encode question and KB contexts independently. In NLP, \cite{bahdanau2014neural} was the first to apply attention model. By cooperatively learning to align and translate, they enhanced the encoder-decoder Neural Machine Translation (NMT) framework. They argued that having source sentence by a fixed vector is illogical, and proposed a soft-align method, which could be understood as attention mechanism. \cite{dong-etal-2015-question} represented questions using three CNNs with different parameters when dealing with different answer contexts including answer path, answer context and answer type. With this inspiration \cite{hao-etal-2017-end} proposed a cross-attention mechanism to encode questions according to various candidate answer aspects. Further more \cite{chen2019bidirectional} proposed bidirectional attention similar to those applied in machine reading comprehension \cite{xiong2016dynamic}\cite{seo2016bidirectional}\cite{wang2016machine} by modeling the interactions between questions and KB contexts. We take a stride forward incorporating BERT\cite{devlin2018bert} pre-trained language model embeddings to encode our question and candidate answer contexts from KB. We fine tuned BERT for the KB-QA problem with a Multi-Head Attention mechanism \cite{DBLP:journals/corr/VaswaniSPUJGKP17} which is based on a Convolution Neural Network (CNN) \cite{kim2014convolutional} encoder. Our proposed architecture is also based on the bi-directional cross attention mechanism between the asked question and KB answer contexts \cite{chen2019bidirectional}.


\section{Overview}

The goal of the KB-QA task can be formalized as follows: given a natural language question q, the model should return an entity set A as answers. The general architecture of a KB-QA system is illustrated in Figure 1. First, the candidate entity of the question is identified, then the candidate answers are generated from the Freebase \cite{bollacker2008freebase} knowledge base. 

The questions are then encoded using pre-trained-Bert model \cite{devlin2018bert} which is fine tuned with a CNN encoder with Multi-Head attention \cite{DBLP:journals/corr/VaswaniSPUJGKP17} for relationship understanding. Then, cross attention neural networks \cite{hao-etal-2017-end} are employed to represent question under the influence of candidate answer set aspects(like entity types, relation paths) and vice versa. Finally, the similarity score between the question and each corresponding candidate answer set is calculated, and the candidates with the highest score will be selected as the final answer. 

\begin{figure}[ht] 
\centerline{\includegraphics[width=8cm, height=9cm]{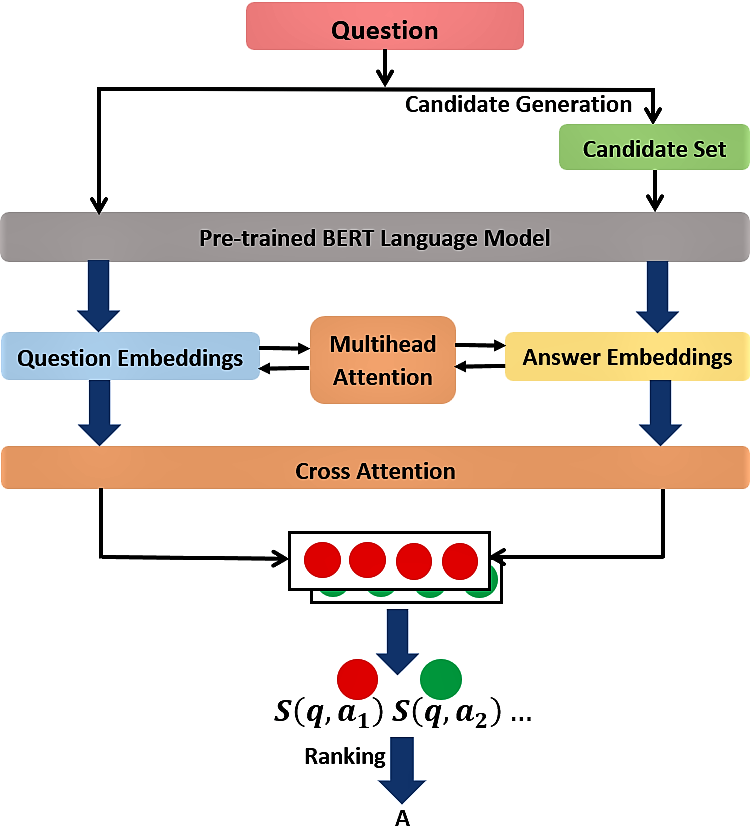}}
\caption{General Architecture of a KB-QA.}
\label{fig}
\end{figure}
 
We refer to Freebase\cite{bollacker2008freebase} as our knowledge base. It has more than 3 billion facts, and is used as the supporting knowledge base for many Question Answering systems. In Freebase, the facts are represented as subject-predicate-object triples (s,p,o). Following is an example triple in Freebase: (/m/01428,  /language/human\textunderscore language/countries\textunderscore spoken\textunderscore in, /m/03\textunderscore r3) which relates to the fact that the language spoken in the country of Jamaica is Jamaican English where "/m/03\textunderscore r3" denotes the country Jamaica while "/m/01428" denotes the language Jamaican English, and  "/language/human\textunderscore language/countries\textunderscore spoken\textunderscore in" is a relationship between the Freebase entities.

\section{Our Approach}
\subsection{Candidate Generation}
Ideally all entities in the Freebase should be considered for candidate answers, but practically, this is computationally expensive and can be avoided. We use Freebase API\cite{bollacker2008freebase} to find a named entity for each question \ q \ which acts as the primary entity for the given question. For general understanding, Freebase API methodology resolved \ 86\% \ of questions in open Web-Questions data-set when top-1 accuracy criteria \cite{yao2014information} is used.

For example if the question "what language Jamaican people speak?" is passed through the Freebase API, it returns "Jamaica" as the main entity. After the named entity is identified with Freebase API, we gather all the other entities that are directly associated to the named entity 
with in 2-hop. These entities create the candidate set.

\subsection{Question Representation}
Initially, we need to collect the representations for each word in the question. These representations contain all the information of the question, and could serve as follows: For example, the question Q is represented as Q = ($x_1$, $x_2$,..., $x_n$) where $x_i$ stands for $i^{th}$ word. The input natural language question Q = $\{ q_i \}_{i=1}^{|Q|} $ as a sequence of word embeddings \(q_i\) fine tuned from Bert pre-trained language model using an encoder layer. 

\subsection{Context Representation}
For each candidate set from the KB, we generate context for each question focusing on three aspects: answer type, answer path, and answer context from KB.

Answer type contains entity type information and helps in narrowing down entities while ranking the answers. If a question uses the word "who", the candidates answers that are relevant to a person are more likely to be correct. Answer path is a sequence of relations from a candidate set to a named entity. For example "/language/human\textunderscore language/countries\textunderscore spoken\textunderscore in" is a relation path between Jamaica and Jamaican English which is stored as [human, language, countries, spoken, in] in the Freebase.

Answer context is defined as the surrounding entities (e.g., neighbour nodes) of the candidate which help to answer the asked questions with constraints. We only consider the context nodes that have overlap with the asked questions. In particular, for each context node (i.e. a sequence of words) of a candidate, we first compute the longest common sub-sequence between the neighbors and the question. If there exists some common sub-sequence between question and neighbor entities we store them as Answer context which will help in answering questions with multiple entities. All this information is stored as the context and will be fed into the pre-trained Bert model to generate embeddings and are fine tuned using an encoder layer.

\subsection{BERT Language Model}
BERT is a multi-layer bidirectional transformer encoder. The given input is a token sequence at the character-level. This means it is either a single sequence or a special token [SEP] separated sentence pair. Each token in the input sequences is the aggregate of the token embeddings, the segment embeddings, and the position embeddings. The initial token in every sequence is consistently a special classification symbol ([CLS]), and for the classification tasks it uses the hidden state of the token. After fine tuning, the pre-trained BERT \cite{devlin2018bert} representations are used in several natural language processing tasks.

\subsubsection{Notation}

For an input sequence of word or sub-word tokens X = (${x_1}$, . . . ,${x_n}$ ), BERT trains an encoder that generates a contextualized representations for each token: $x_1$, . . . , $x_n$ = enc(${x_1}$, . . . , ${x_n}$). Each token in the sequence, positional embeddings $p_1$, . . . , $p_n$ are used to label the absolute position of input sequence because deep transformer is used to implement the encoder.

\subsubsection{Masked Language Modeling (MLM)}
Masked Language Modeling (MLM) or "Cloze test", is used to predict the missing tokens from their placeholders in a given sequence. When a subset of tokens Y $\subseteq$ X sampled and substituted with placeholder set of tokens.In Bert's MLM implementation ,Y accounts for 15\% of the tokens in X; of those, 80\% are replaced with [MASK], 10\% are replaced with a random token (according to the unigram distribution), and 10\% are kept unchanged. The main idea is to predict the modified input from the original tokens in Y. BERT selects each token in Y independently by randomly selecting a subset.

\subsubsection{Next Sentence Prediction (NSP)}
The Next Sentence Prediction (NSP) predicts whether $X_B$ is immediate continuation of $X_A$ when two sequences $X_A$ and $X_B$ are given as input. In BERT methodology, it initially takes $X_A$ from corpus. Then it either reads $X_B$ from where $X_A$ has terminated or it randomly samples $X_B$ from different point in corpus. A special token [SEP] is used to separate the two sequences. Also, a special token [CLS] is appended to $X_A$ and $X_B$ to express as an input. The main idea of [CLS] is to truly find whether $X_B$ follows $X_A$ in the corpus.

\subsection{Embedding Layer}

To embed question and KB context we use pre-trained BERT\cite{devlin2018bert} language model. Without the need of Recurrent Architecture BERT uses positional embeddings to encode the word sequences. By using unsupervised learning based on Masked Language Models (MLM) and Next Sentence Prediction (NSP), BERT embeddings are generated. These embeddings contain bidirectional attention from Masked Language Model(MLM) section of BERT.

BERT model takes sentences in one or two formats. If we have a context of [location, language, human, language, countries,spoken, in, Jamaican] and a question [what does Jamaican people speak?], we can either encode it as a single input or two separate inputs to the BERT Model as shown below.

<CLS>[question]<SEP> [context] Or 

<CLS>[question] 

<CLS>[context]

\begin{figure*}[ht]
\includegraphics[width=1\textwidth]{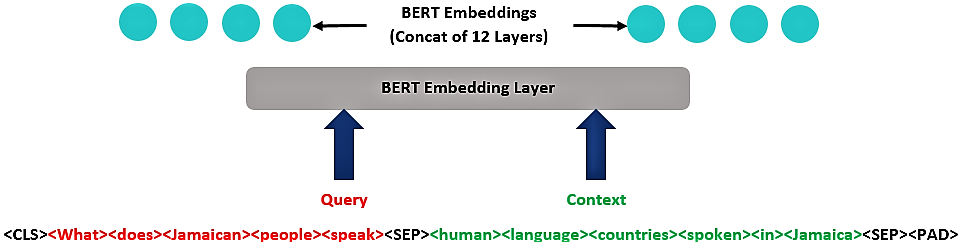}
\caption{BERT Embedding Layer}
\label{fig}
\end{figure*}

We provide Context and Query together with query first followed by the context as shown in the figure above. Since BERT is trained on "next sentence prediction" outcome, we believe that using this formulation will provide richer query and context embeddings. The maximum length of the context and the question are 96 and 18 respectively. We use the concatenation of all 12 layers of BERT model as our embedding layer.

\subsection{Encoder Layer}
The encoder layer is a stack of following layers:

[CNN-layer + self-attention-layer + feed-forward-layer]

Similar to \cite{seo2016bidirectional}, the input of this layer at each position is [c, a, c $\bigodot$ a, c $\bigodot$ b], where a and b are respectively a row of attention matrix A and B. For the self-attention-layer, we adopted multi-head attention mechanism defined in \cite{DBLP:journals/corr/VaswaniSPUJGKP17} which is the improvement of the basic attention mechanism.
           \begin{equation}
                {head_i} = Attention(QW_i^Q, KW_i^K, VW_i^V)
            \end{equation}\begin{equation}
                Multi - Head(Q,K,V) = Concat(head_1,\cdot\cdot\cdot\cdot\cdot, head_h)
            \end{equation}\begin{equation}
               SelfMulti-Head = Multi - head(X, X, X)
           \end{equation}

These three equations are the formation process of the self multi-head attention mechanism. The matrix of W is the weight matrix. The Q, V, K should multiply its corresponding weight matrix before getting into the attention function. Repeat this process h(number of heads) times and connect each result then we can get a new vector matrix that reflects the relationship between Q, V. Especially, in the self multi-head attention mechanism, to look for internal connections within words, the Q = V = K = X, and X represents the word vector matrix.

The multi-head attention mechanism helps the model learn the words relevant information in different presentation sub-spaces. The self-attention mechanism can extract the dependence in words. As the name shows, the self multi-head attention mechanism integrates the benefits of both, creates a context vector for each word. Then we don't need to depend on additional information to get a matrix that reflects the context relationship between current word and other words in a sequence. Each of these basic operations (cnn/self-attention/ffn) is placed inside a residual block

\subsection{Attention Layer}
We use a focused, Context-Query attention layer on top of the pre-trained BERT embeddings identical to that of the QANet\cite{yu2018qanet} model. Such attention modules were standard in other KB-QA models such as \cite{hao-etal-2017-end}\cite{chen2019bidirectional}.

C is used to denote context from KB and Q for natural language question. The context-to-query attention is constructed as follows:(1) Compute similarities between each pair of context and query words that generates a similarity matrix S $\in R^{n*m}$. (2) Normalize each row of S by applying soft-max function. (3) Compute context-to-query attention A = S$\cdot$ QT$ \in R^{n*d}$

The similarity function used here is the tri-linear function as mentioned in \cite{seo2016bidirectional} :
f(q, c) = $W_0$[q, c, q $\bigodot$ c]

\section{Experiments}
\subsection{Dataset}\label{AA}
We use the Web-Questions\cite{berant-etal-2013-semantic} dataset. Web-Questions dataset is built by using the Google Suggest API to obtain questions that begin with a wh-word and contain exactly one entity. Specifically, they queried the question excluding the entity, the phrase before the entity, or the phrase after it. Each query generates 5 candidate questions, which are added to the queue. Further they iterated until 

1 million questions were visited; a random 100K were submitted to Amazon Mechanical Turk (AMT).
The AMT task requested that the distributed workforce should answer the question using only the Freebase page of the questions' entity, or otherwise mark it as unanswerable by Freebase. The answer was restricted to be one of the possible entities, values, or list of the entities on the page. Thus,
this combination of Web-Questions dataset with Freebase KB is used as the base line.

\subsection{Comparison with Sate of the art}

To assess the proposed methodology, experiments were conducted on the Freebase KB and the Web-Questions \cite{berant-etal-2013-semantic} dataset. Freebase is large-scale KB\cite{bollacker2008freebase} that has organised general facts as subject-predicate-object triples. It has 41M non-numeric entities with 19K unique properties and 596M assertions.The Web-Questions\cite{berant-etal-2013-semantic} dataset has 3,778 question-answer pairs for training and 2,032 for testing. The questions are gathered from Google Suggest API, and the answers from Amazon Mechanical Turk which are labelled manually. All of the answers are from the Freebase KB. we use 80\% of the training data as training set and 20\% as validate set. The evaluation metric we use is F1 score, and the average result is calculated by \cite{berant-etal-2013-semantic} script.

\subsubsection{Settings}

For KB-QA training, we use pre-trained Bert embeddings base uncased version. During tokenization, BERT\cite{devlin2018bert} code uses a word-piece algorithm to split words into sub-words and all less frequent words will be split into two or more sub-words. The vocab size of Bert is 30522. We adopt delexicalization strategy as mentioned in \cite{chen2019bidirectional}. For each question, the candidate entity mentions those belonging to date, ordinal, or number are replaced with their type. Same is applied on answer context from KB text, if the overlap belongs to above type. This assures that the query matches up with answer context in the embedding space. The dropout rates for both question and answer encoder side is set to 0.3. The bath size is set as 4 and answer module threshold is set to 0.7 to allow multiple answers for questions with list of answers. We use Adam optimizer\cite{kingma2014adam} to train the model. The initial learning rate is set to 0.01. Further learning rate is reduced by a factor of 10 if no improvement is observed in validation process for 3 successive epochs. The hyper-parameters are tuned on validation-set.

\subsubsection{Results}

In this section, we compare the performance of our method with other IR-based approaches. The results are shown in Table 1. Based on this table our method (LMKB-QA) obtained an F1 score of 52.7 on Web-Questions using the topic entity predicted by Freebase API. As Table 1 shows our method achieves better results or even competes with state-of-the-arts. This demonstrate the effectiveness of our idea to use Bert pre-trained language model embeddings in the Question Answering on Knowledge Base problem.

\begin{table}[ht]
\def\tablename{Table}
\centering
    \begin{tabular}{|p{5cm}|p{1.5cm}|} 
    \hline
    \textbf{Methods} & \textbf{Avg F1} \\ [0.5ex]
    \hline
    Bordes 2014b\cite{bordes2014open} &29.7 \\
    Bordes 2014a\cite{bordes2014question} &39.2 \\
    Yang 2014\cite{yang-etal-2014-joint} &41.3 \\
    Dong 2015\cite{dong-etal-2015-question} &40.8 \\
    Bordes 2015\cite{bordes2015large} &42.2 \\
    Xu 2016 \cite{xu-etal-2016-question}&42.2 \\
    Hao 2017\cite{hao-etal-2017-end} &42.2 \\
    Chen 2019\cite{chen2019bidirectional}&49.7\\[1ex]
    Chen 2019\cite{chen2019bidirectional} topic entity predictor &51.8\\[1ex]
    \hline
    Our Approach &52.7\\[1ex]
    \hline
    \end{tabular}
    \caption{Evaluation results on Web-Questions}
    \label{table:default}
\end{table}

It is important to note that our proposed approach is based on a CNN architecture, and only depends on the training data (it does not depends on wiki text \cite{xu-etal-2016-hybrid}). \cite{bordes2014open} applies BOW method to obtain a single vector for both questions and answers. \cite{bordes2014question} further improve their work by proposing the concept of sub-graph embeddings. Besides the answer path, the sub-graph contains all the entities and relations connected to the answer entity. The final vector is also obtained by bag-of-words strategy. \cite{yang-etal-2014-joint} follows the SP-based manner, but uses embeddings to map entities and relations into KB resources, then the question can be converted into logical forms. \cite{dong-etal-2015-question} use three columns of Convolution Neural Networks (CNNs) to represent questions corresponding to three aspects of the answers, namely the answer context, the answer path and the answer type. \cite{bordes2015large} put KB-QA into the memory networks framework \cite{sukhbaatar2015end}, and achieves the state-of-the-art performance of end-to-end methods. Our approach incorporates pre-trained Bert language model embeddings fine tuned for the KB-QA task using a Encoder architecture with multi head attention. \cite{bordes2014question,bordes2015large,bordes2014open} all utilize BOW model to represent the questions, while ours takes advantage of pre-trained language model embeddings. Also note that \cite{bordes2015large} uses additional training data such as Reverb \cite{fader-etal-2011-identifying} and their original data set Simple-Questions.

The proposed method in \cite{dong-etal-2015-question} employs three fixed CNNs to represent questions. \cite{hao-etal-2017-end} implemented the mutual influence between the representation of questions and the corresponding answer aspects. \cite{chen2019bidirectional} further enhanced the result by using memory networks \cite{weston2014memory} to control the mutual influence between the representation of questions and the corresponding answer aspects. The approach in \cite{chen2019bidirectional} achieved an F1 score of 0.518 using a custom topic entity predictor while an F1 score of 0.497 has achieved via the Freebase Search API.  The \cite{yih-etal-2016-value, yih-etal-2015-semantic} have higher F1 scores than other methods. Their approach was able to address more questions with constraints and aggregations. But, their methods apply number of manually designed rules and features, which come from the observations on the training questions set. This is a manual process which reduces the versatility of their proposed approach. Integrated systems like \cite{xu-etal-2016-hybrid,xu-etal-2016-question} generates higher F1 score leveraging Wikipedia free text as external knowledge, which are based on semantic parsing. Therefore the systems are not directly compared to ours.

\section{Conclusion}
In this paper, we focused on using a pre-trained language model for KB-QA task. Firstly, we used \textit {Bert base uncased} for the initial experiments. We further fine tuned these embeddings with two way attention mechanism from the knowledge base to the asked question and from the asked question to the knowledge base answer aspects. Our method is based on a simple CNN architecture with Multi-Head Attention mechanism to represent the asked question dynamically in the multiple aspects. Our experimental results show the effectiveness of the Bert pre-trained language model embeddings. 

In the future, we would like to explore other language models including GPT-2, RoBERTa, Transformer-XL and XLNet. We will also further investigate to answer even more complex questions on the knowledge bases. 

\section{Acknowledgements}

We would like to thank Hugging Face\cite{Wolf2019HuggingFacesTS} for their open-source code to use pretrained BERT embeddings for downstream tasks using PyTorch, BAMnet\cite{chen2019bidirectional} for open-source code to process Freebase knowledge base dump,  and Web-Questions\cite{berant-etal-2013-semantic} for the train, test dataset, and F1 metric script for comparison. 

\medskip
\bibliographystyle{plain}
\bibliography{refer}{}
\end{document}